\documentclass[10pt,twocolumn,letterpaper]{article}

\usepackage{cvpr}              

\usepackage{graphicx}
\usepackage{amsmath}
\usepackage{amssymb}
\usepackage{booktabs}
\usepackage{multirow}
\usepackage[title]{appendix}

\usepackage[pagebackref,breaklinks,colorlinks]{hyperref}

\usepackage[capitalize]{cleveref}
\crefname{section}{Sec.}{Secs.}
\Crefname{section}{Section}{Sections}
\Crefname{table}{Table}{Tables}
\crefname{table}{Tab.}{Tabs.}

\def\cvprPaperID{2344} 
\def\confName{CVPR}
\def\confYear{2022}

\begin{document}

\title{Day-to-Night Image Synthesis for Training Nighttime Neural ISPs}

\author{Abhijith Punnappurath$^1$ \and Abdullah Abuolaim$^2$\thanks{Work done while an intern at the Samsung AI Center -- Toronto.} \and Abdelrahman Abdelhamed$^1$ \and Alex Levinshtein$^1$ \quad \quad \quad Michael S. Brown$^{1}$\\
$^1$Samsung AI Center -- Toronto \quad \quad \quad \quad $^2$York University\\
{\tt\small \{abhijith.p,a.abdelhamed,alex.lev,michael.b1\}@samsung.com, abuolaim@eecs.yorku.ca}
}
\maketitle

\begin{abstract}
Many flagship smartphone cameras now use a dedicated neural image signal processor (ISP) to render noisy raw sensor images to the final processed output.   Training nightmode ISP networks relies on large-scale datasets of image pairs with: (1) a noisy raw image captured with a short exposure  and a high ISO gain; and (2) a ground truth low-noise raw image captured with a long exposure and low ISO that has been rendered through the ISP.  Capturing such image pairs is tedious and time-consuming, requiring careful setup to ensure alignment between the image pairs. In addition, ground truth images are often prone to motion blur due to the long exposure.  To address this problem, we propose a method that synthesizes nighttime images from daytime images.   Daytime images are easy to capture, exhibit low-noise (even on smartphone cameras) and rarely suffer from motion blur.   We outline a processing framework to convert daytime raw images to have the appearance of realistic nighttime raw images with different levels of noise.  Our procedure allows us to easily produce aligned noisy and clean nighttime image pairs.  We show the effectiveness of our synthesis framework by training neural ISPs for nightmode rendering. Furthermore, we demonstrate that using our synthetic nighttime images together with small amounts of real data (e.g., $5\%$ to $10\%$) yields performance almost on par with training exclusively on real nighttime images. Our dataset and code are available at \url{https://github.com/SamsungLabs/day-to-night}.
\end{abstract}

\section{Introduction}
\label{sec:intro}

\begin{figure}[!t]
\centering
\includegraphics[width=1.0\linewidth]{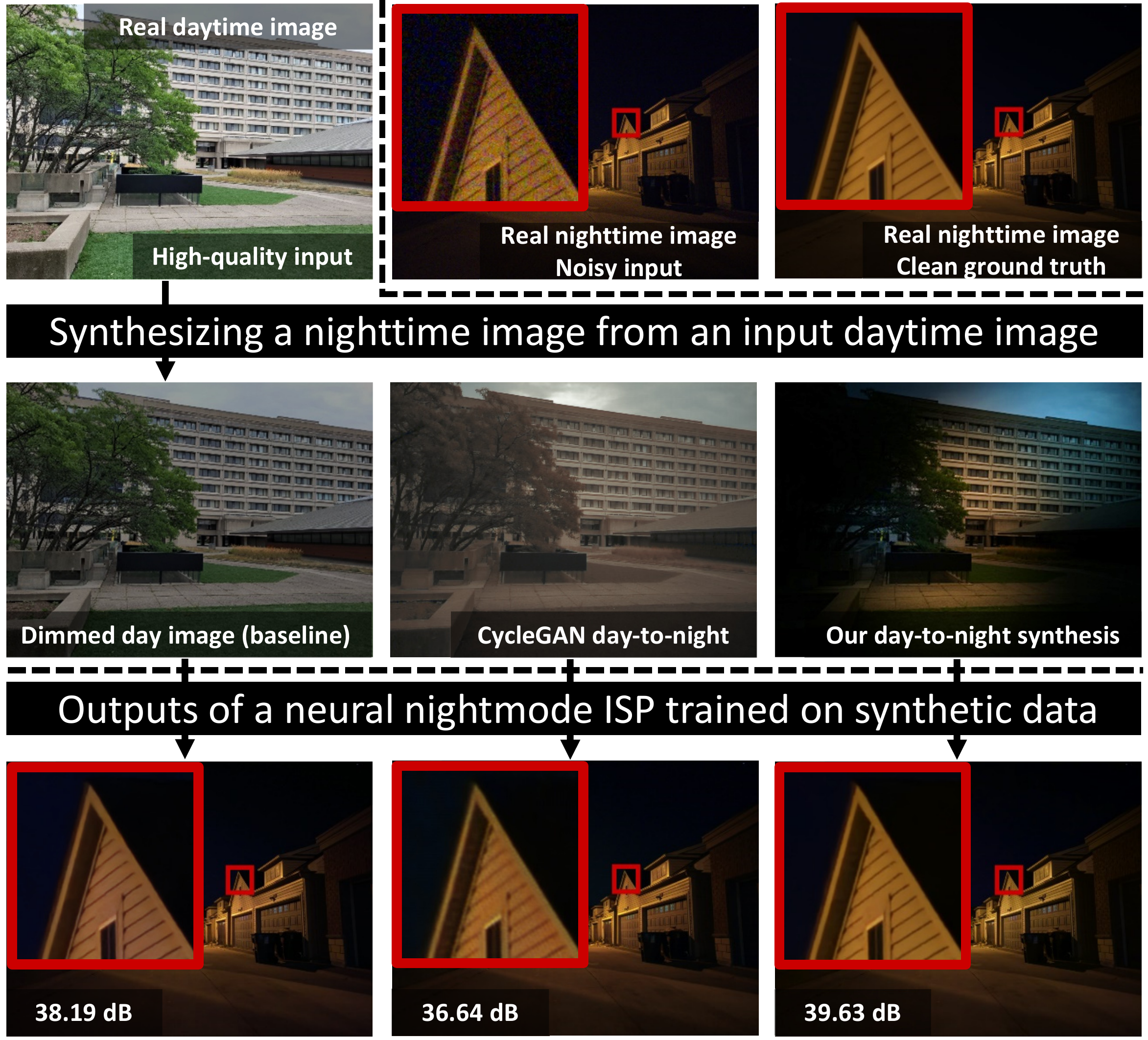}
\caption{Top row: A high-quality daytime image and a noisy/clean nighttime image pair. While high-quality day images are easy to capture, paired nighttime data, needed for training nightmode neural ISPs, is challenging to acquire. We propose a day-to-night image synthesis framework that converts the daytime images to appear as noisy/clean paired nighttime images suitable for training DNNs. Middle row: Different day-to-night image synthesis methods. Bottom row: A neural nightmode ISP trained by our synthetic night images produces more accurate results compared to models trained on dimmed day images, or night images synthesized by CycleGAN~\cite{cyclegan}. Inset shows PSNR in dB.
\label{fig:teaser}}
\end{figure}

Capturing images at night and in low-light environments is challenging due to the low photon count hitting the camera sensor.  Because of the weak signal, the image must be gained (i.e., high ISO), which further amplifies the sensor noise.    This is particularly troublesome for smartphone cameras, where the sensor's small form factor limits the amount of light per pixel photosite, resulting in significant noise levels in low-light and night environments. When noisy sensor images are processed by the camera's image signal processor (ISP), the noise is often amplified, resulting in noisy and aesthetically unappealing final standard RGB (sRGB) output images.  One solution is to capture the scene using a long exposure (e.g., several seconds), but this often is not viable as it requires the camera to be placed on a tripod to avoid camera shake and the scene needs to remain static to avoid motion blur.  Recent advancements in deep networks designed to render noisy raw nighttime images to the processed sRGB outputs have shown impressive results. These networks are trained on aligned noisy/clean image pairs captured in low-light and nighttime environments.

Several recent works have undertaken the effort to capture high-quality ground-truth images of low-light scenes at night, such as~\cite{sid, chen2019seeing, sidd, wang2019underexposed}.  These training images are collected using two strategies. The first is to capture a noisy short-exposure high-ISO gain image as input, together with a long-exposure low-ISO gain image to serve as the target ground truth (e.g., \cite{sid, chen2019seeing}).  Images must be carefully captured on a tripod to ensure alignment.  This approach is prone to blurring in the ground truth image due to the long exposure, especially when capturing outdoor scenes, where it is harder to control motion in the environment.   The second strategy is to capture multiple short-exposure images with high ISO.  The ground truth image is computed by averaging the image sequence to reduce noise (e.g., ~\cite{sidd}).  This method has its own challenges in terms of alignment and image sequence fusion. In addition, any motion in the scene will result in motion blur in the averaged ground truth image. Further confounding data collection is the fact that data capture and network training are required {\it per} sensor, as the raw images are sensor specific~\cite{afifi2019SIIE, two-cam-wb}.

From a practical standpoint, capturing data for night mode represents a significant burden.  This is especially true for smartphone cameras, where sensors are being continually updated and many devices now have multiple cameras per device with different underlying sensors~\cite{two-cam-wb}.  Recent approaches have tried to overcome the need for nighttime data collection by using deep neural networks (DNNs) to synthesize them, such as~\cite{anoosheh2019night, isola2017image, yi2017dualgan, sun2019see}. However, such methods require large datasets for training and tend to produce artifacts and sometimes unrealistic image content.

\noindent \textbf{Contribution}~We present a method to reduce the reliance on carefully captured paired nighttime images. Specifically, we propose a procedure that processes daytime images to produce pairs of high-quality and low-quality nighttime images as shown in Fig.~\ref{fig:teaser}. Our approach incurs minimal loss in image details and structure and does not require large datasets of nighttime images. Unlike capturing nighttime images under low-light conditions, capturing daytime images with proper exposure is straightforward and does not require careful camera or scene setup.  We show that our proposed framework is useful for nighttime image processing and enhancement through training neural ISPs to render noisy night raw-RGB images to their final clean sRGB outputs. We demonstrate that training on our synthetic nighttime images mixed with a small amount of real data (e.g., $5\%$ to $10\%$) produces performance almost on par with training exclusively on real nighttime images. Our method significantly reduces the time and effort needed to deploy neural ISPs targeting nighttime imaging.

\section{Related work}

In this section, we use the standard naming convention of referring to minimally processed sensor images as {\it raw-RGB} images, and images that have been processed by the camera's ISP as {\it sRGB} images.  Our paper is focused on synthesizing nighttime raw-RGB images to train an in-camera neural ISP that takes a raw-RGB image as input and outputs a processed sRGB image.

\noindent\textbf{Paired nighttime images}~A critical component in developing a DNN-based nightmode neural ISP is preparation of training data in the form of paired long- and short-exposure nighttime images. The SID dataset~\cite{sid} includes captured pairs of long-exposure sRGB and short-exposure raw-RGB images used for training a convolutional neural network (CNN) pipeline for nighttime and low-light image processing. Similarly, the work in~\cite{chen2019seeing} involved capturing paired long- and short-exposure static raw-RGB videos, used for video processing and enhancement. With focus on image enhancement, \cite{wang2019underexposed} collected 3000 underexposed image pairs, containing many nighttime images, that were processed by expert photographers to produce corresponding ground-truth high-quality images. Nighttime images are still inherently low-quality due to low light and high noise, which lead to low signal-to-noise ratio (SNR). Alternative approaches involve the use of bursts or sequences of images~\cite{hasinoff2016burst, liu2014fast, godard2018deep, sidd} to produce high-quality nighttime images. However, burst alignment algorithms may fail in extreme low-light conditions, are blur-prone, and produce alignment artifacts.

\noindent\textbf{Day-to-night image synthesis}~Closely related to our approach are recent methods that synthesize nighttime images given high-quality daytime images. Generative adversarial networks (GANs), such as CycleGAN~\cite{cyclegan} and EnlightenGAN~\cite{jiang2021enlightengan}, play the major role in this category. In~\cite{isola2017image}, a conditional GAN (cGAN) is used for image-to-image translation from daytime to nighttime images. DualGAN~\cite{yi2017dualgan} is proposed for both day-to-night and night-to-day image translation. The method in~\cite{laffont2014transient} proposed editing an outdoor scenes dataset to synthesize outdoor images with general attributes such as ``night'', ``dusk'', and ``fog'', which can be used for training day-to-night image-to-image translation GANs. The method in~\cite{sun2019see} proposed a nighttime image semantic segmentation framework using GANs to convert between nighttime and daytime images to improve segmentation. Cross-domain car detection using GANs for unsupervised day-to-night image translation is attempted in~\cite{arruda2019cross}, while ~\cite{romera2019bridging} transfers annotated daytime images to nighttime so that the annotations can be reused through data augmentation. Most of such methods do not target nighttime image processing directly, and they tend to produce low-quality or unrealistic nighttime images that may contain hallucinated content. Furthermore, most of such methods target synthetizing sRGB images and not raw-RGB images.

\noindent\textbf{Night-to-day image synthesis}
Instead of converting daytime images to nighttime ones, some methods\cite{lin2020gan,anoosheh2019night,zheng2020forkgan} propose to enhance the nighttime images by converting them to daytime ones, which may improve the model training for target tasks such as nighttime vehicle detection, segmentation, visual localization etc. Generally, night-to-day image translation may work well as a form of data preprocessing or augmentation; however, due to the inherent low-quality of nighttime images, such methods cannot be used for synthesizing high-quality paired nighttime images.

\noindent\textbf{Domain adaptation}~Recent methods perform domain adaptation to close the gap between model performance on nighttime images versus daytime images. The method in~\cite{lengyel2021zero} performs domain adaptation using a physics prior to reduce the day-vs-night distribution shift in neural network feature maps. In~\cite{schutera2020night}, domain adaptation is performed via online image-to-image translation from day to night images.

\noindent\textbf{Special care for nighttime images}
Another line of work takes special care when processing nighttime images. For example, the method in~\cite{ancuti2020day} performs nighttime image dehazing with special treatment of multiple localized artificial light sources usually present in nighttime images. Similarly,~\cite{santra2016day} handles spatially varying atmospheric light that may be present in nighttime images. Also, the method in~\cite{yu2010traffic} takes special care of night conditions while performing traffic light detection.

Unlike all discussed methods, our method targets the synthesis of paired high- and low-quality raw-RGB nighttime images from daytime images without the need for large datasets or training large GAN models. As we show in our experiments, our synthesized paired images are more suitable for training nightmode neural ISPs.


\section{Nighttime image synthesis}
\label{sec:method}
Our synthesis procedure is applied to the raw Bayer image, which has not been demosaiced (i.e., there is one channel per pixel defined by the Bayer pattern). Our procedure involves removing the illumination in the day image, lowering the exposure, relighting the scene with night illuminants, and adding noise to mimic a real nighttime image.

\begin{figure*}[!t]
\centering
\includegraphics[width=1.0\linewidth]{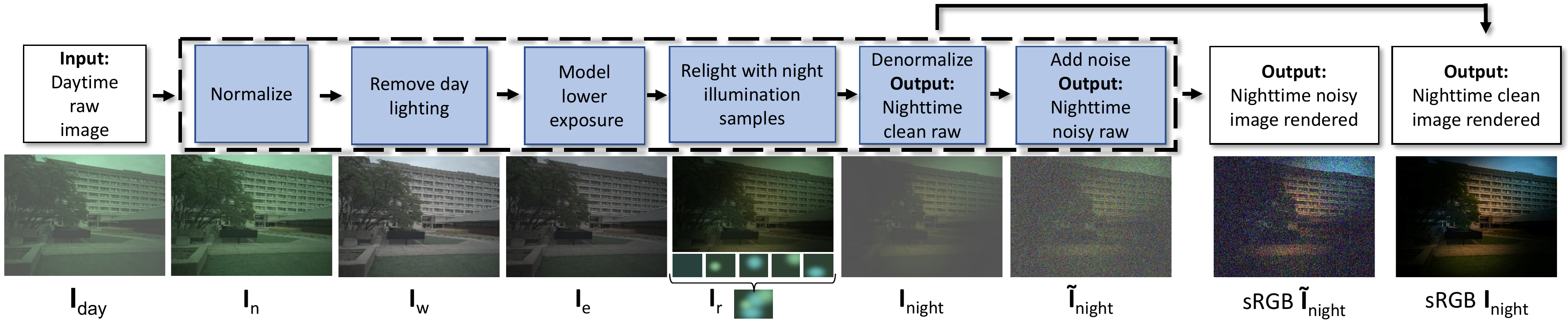}
\caption{An overview of our proposed day-to-night image synthesis framework. Our procedure involves removing illumination in the day raw image, lowering the exposure, relighting the scene with night illuminants, and adding noise to mimic a real nighttime raw image. For visualization, the raw images have been demosaiced, and a gamma has been applied.}
\label{fig:method}
\end{figure*}

An overview of our proposed synthesis framework is shown in Fig.~\ref{fig:method}. We assume that the input daylight images are sharp and noise-free. This is generally a valid assumption because well-lit outdoor scenes can be captured with a short exposure and low ISO, and good SNR properties~\cite{sidd}.

Our pipeline begins with the minimally processed Bayer image recorded by the camera sensor. Let us denote the stacked RGGB Bayer image by $\mathbf{I}_{\mbox{\scriptsize{day}}} \in \mathbb{R}^{ \frac{H}{2} \times \frac{W}{2} \times 4}$, where $H$, $W$ denote the image size in pixels. We first adjust the black- and white-levels, and normalize the data. The normalized image $\mathbf{I}_n = (\mathbf{I}_{\mbox{\scriptsize{day}}} - b_l) / (w_l - b_l)$, where $b_l$ and $w_l$ denote the black- and white-level given by the camera's metadata. Cameras typically apply a white-balance routine to the raw image to remove the color cast introduced by the scene illumination. This is a two-step procedure that involves estimating the RGB values corresponding to the scene illuminant, and dividing out each RGB color channel by the corresponding value to compensate for the illumination. While this process ensures that the achromatic colors are corrected, it cannot guarantee that all the colors are illumination-compensated. Compared to other illuminations, images captured under daylight have the special property that they incur the least error in terms of overall color correction when white-balanced~\cite{cheng2015beyond}. To apply our synthesis model, we specifically select images captured outdoor, under daylight illumination. We remove the day lighting in the image by applying the white balance, as the next step of our pipeline. The white-balanced image $\mathbf{I}_w = \mathbf{I}_n \mathbf{L}_{\mbox{\scriptsize{day}}}$, where $\mathbf{L}_{\mbox{\scriptsize{day}}} = \mbox{diag}(\frac{1}{r}, \frac{1}{g}, \frac{1}{g}, \frac{1}{b})$, where we use the day illuminant estimate from the camera's auto-white-balance (AWB) routine to construct $\mathbf{L}_{\mbox{\scriptsize{day}}}$. The green-channel values $g$ in
$\mathbf{L}_{\mbox{\scriptsize{day}}}$ are typically normalized to 1.

Nighttime images usually have a lower average brightness than daytime images.  In addition, nighttime images are often illuminated by multiple light sources with different spectral properties. The next two stages of our pipeline are designed to model these effects. We pre-compute a dictionary $\mathcal{D}$ of average nighttime brightness values, and a dictionary $\mathcal{L}$ of the appearances of nighttime illuminants using a small set of real nighttime images.

We construct $\mathcal{D}$ by computing the normalized mean intensity value $d$ of each Bayer image. To build the nighttime illuminant dictionary $\mathcal{L}$, we image gray cards under different commonly encountered night illuminations. Next, we lower the exposure of $\mathbf{I}_w$ by multiplying the image by a global scale factor $d$. The resulting dimmed image is $\mathbf{I}_e = \mathbf{I}_w * d$. This scale factor $d$ is randomly sampled around a distribution constructed from our dictionary $\mathcal{D}$ of average nighttime brightness.

To relight the scene, we randomly sample a small set (usually five to seven) of nighttime illuminants. First, we fit a 2D multivariate Gaussian distribution of joint chromaticity values ($\frac{r}{g}$ and $\frac{b}{g}$) around our database of night illuminations $\mathcal{L}$. Then, we randomly sample night illuminant $\mathbf{y}$ from this distribution as follows:
\begin{equation}
\mathbf{y}
\thicksim
\mathcal{N} \left( \mathbf{\mu}, \mathbf{\Sigma} \right) \ ,
\label{eqn1}
\end{equation}
\begin{equation}
\mathbf{\Sigma}=\frac{1}{M}\sum_{i=1}^M\left( ([\frac{r}{g},\frac{b}{g}]_i-\mathbf{\mu})^\intercal ([\frac{r}{g},\frac{b}{g}]_i-\mathbf{\mu}) \right)\ ,
\label{eqn2}
\end{equation}
where $\mathbf{\mu}$ and $\mathbf{\Sigma}$ are the mean and covariance of the normalized chromaticity values in $\mathcal{L}$, respectively, $M$ is the number of night illuminants in $\mathcal{L}$, and $\mathbf{y}, \mathbf{\mu} \in \mathbb{R}^2$ and $\mathbf{\Sigma}\in \mathbb{R}^{2\times 2}$. See Section~\ref{sec:illum} of supplementary for examples.
We re-illuminate the scene using these night lights with their location in the image and falloff modeled using 2D Gaussian functions with random centering and standard deviation. In particular, the relit image $\mathbf{I}_r$ can be expressed as:
\begin{eqnarray}
\mathbf{I}_r =  \frac{ \sum_{i=1}^{N} \mathbf{I}_e  \mathbf{L}_{{\mbox{\scriptsize{night}}}_i} \odot w_i  \mathbf{M}_i } { \sum_{i=1}^{N} w_i  \mathbf{M}_i }.
\end{eqnarray}
Here $\mathbf{L}_{{\mbox{\scriptsize{night}}}_i} = \mbox{diag}(r_i,g_i,g_i,b_i)$, with $i=\{1,\dots,N\}$, represents the set of night illumination samples. The scalar $w_i$ is used to control the strength of the light source. The mask $\mathbf{M}_i$ is modeled as a 2D Gaussian function $G(x_i,y_i,\sigma_{x_i},\sigma_{y_i})$. We randomly position the light source with center $(x_i,y_i)$ that lies within the image excluding a $10\%$ boundary. The spread of the light source is modulated by $(\sigma_{x_i},\sigma_{y_i})$ which we randomly choose to lie between $[0.5, 1]$ of the image size. The same Gaussian kernel is applied to all channels -- namely $\mathbf{M}_i \in \mathbb{R}^{ \frac{H}{2} \times \frac{W}{2} \times 4}$. The operator $\odot$ denotes element-wise multiplication. Additionally, we select one of the illuminants, $i=1$, as an ambient light, with $\mathbf{M}_1$ being a mask of all 1s, and having a weak strength $w_1$ set between $5\%$ to $10\%$ of the other illuminants. We finally denormalize $\mathbf{I}_r$ to obtain our synthetic nighttime image $\mathbf{I}_{\mbox{\scriptsize{night}}} = \mathbf{I}_r (w_l - b_l) + b_l$.

The image $\mathbf{I}_{\mbox{\scriptsize{night}}}$ at this stage represents a high-quality long-exposure low-ISO nighttime image. This synthesized raw image can now be rendered through the image signal processor to produce the final display-referred sRGB image, which is usually used as the target while training a DNN. Adding noise to the modified raw image $\mathbf{I}_{\mbox{\scriptsize{night}}}$ yields the low-quality short-exposure high-ISO image, which is often the input to the DNN. We adopt the well-established heteroscedastic Gaussian model~\cite{Foi2009ClippedDenoising, Foi2015PracticalRaw-data, Makitalo2013OptimalNoise, Liu2014PracticalImage} for noise. The noisy raw image is generated as follows:
\newcommand{\imgnight}{\mathbf{I}_{\mbox{\scriptsize{night}}}}
\newcommand{\imgnightnoisy}{\tilde{\mathbf{I}}_{\mbox{\scriptsize{night}}}}
\begin{equation}
\imgnightnoisy
\leftarrow
\imgnight +
\mathcal{N} \left( \mathbf{0}, \beta_1 \imgnight + \beta_2 \right) \ ,
\end{equation}
where $\beta_1$ and $\beta_2$ are the shot and read noise parameters.
This noisy raw image $\imgnightnoisy$ when rendered through the ISP closely resembles a typical low-quality short-exposure high-ISO nighttime photo. We empirically determine the values of $\beta_1$ and $\beta_2$ for different ISO levels based on measuring the noise of real noisy/clean nighttime image pairs.


\section{Dataset}

\begin{figure*}[!t]
\centering
\includegraphics[width=1.0\linewidth]{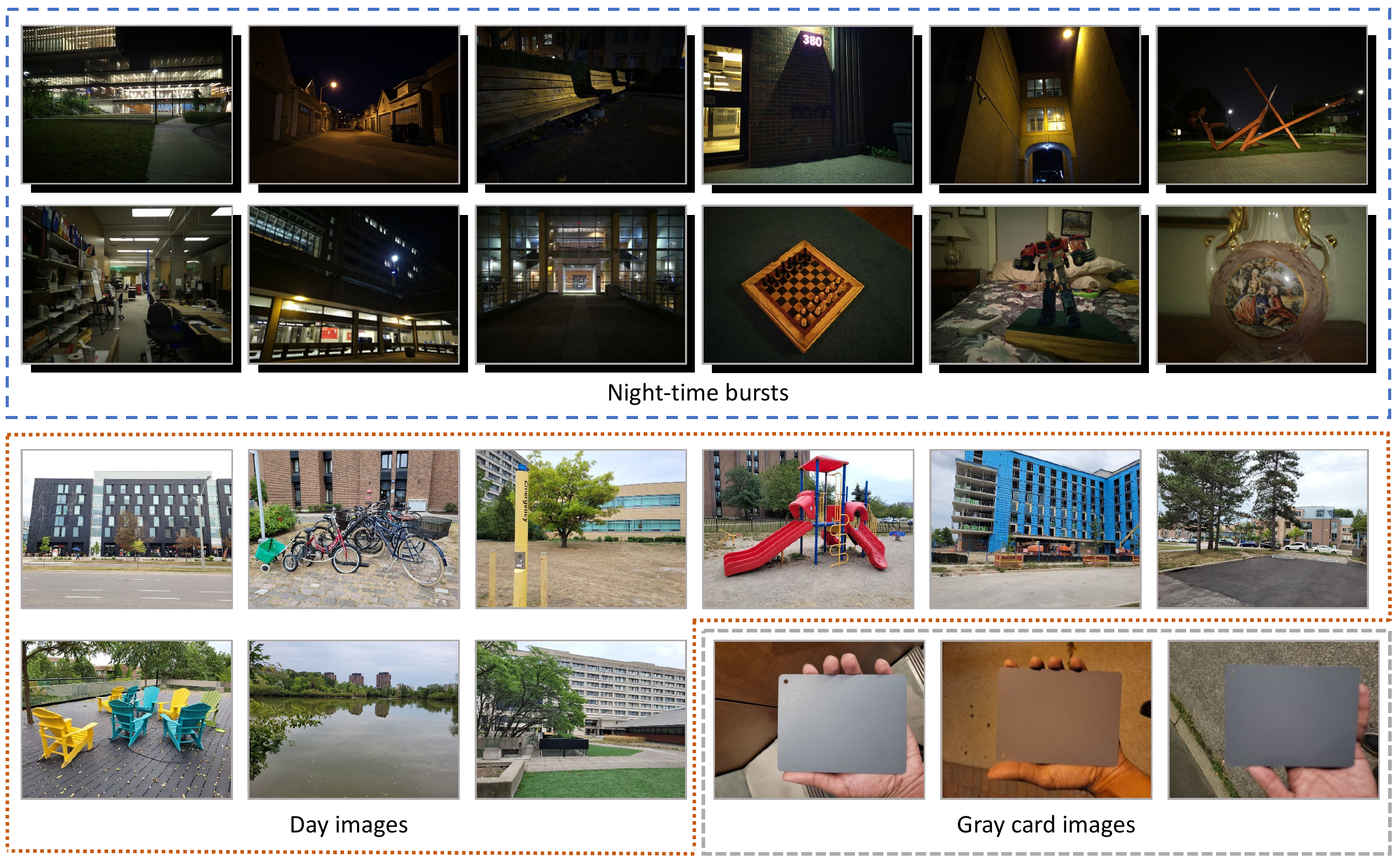}
\caption{Representative examples from our dataset. Our dataset contains nighttime bursts, day images, and images of a gray card.}
\label{fig:dataset}
\end{figure*}

We evaluate our algorithm by using the synthetic data generated by it to train a  nighttime neural ISP. In particular, we investigate two scenarios: (1) the input nighttime raw image is noise-free, allowing us to examine the color rendering accuracy in isolation, and (2) the input raw-RGB data is noisy, which is the more realistic case of an actual nighttime neural ISP. The first task is aimed at demonstrating the ability of our algorithm to closely mimic the illumination in a real nighttime image. Thus, we assume a noise-free input since our emphasis is on color rendering alone. We show that a network trained to render our \emph{synthetic} nighttime input raw image to sRGB performs almost on par with a rendering network trained exclusively on \emph{real} nighttime data. We further demonstrate the utility of our approach as a data augmentation strategy. Adding a small amount ($5\%$ to $10\%$) of real data to our synthetic images allows us to close the performance gap to the purely-real-data model. For the second task, we focus on a full-fledged nighttime neural ISP that renders a noisy short-exposure high-ISO raw image to its corresponding noise-free long-exposure low-ISO processed sRGB image. Similar to the first task, we show that a network trained on our synthetic nighttime raw images with synthetically added noise, and augmented with a minimal amount of real data, achieves performance close to that offered by training only on real nighttime noisy/clean paired data. Note that in both scenarios, the network takes a single demosaiced linear raw frame as input and outputs the processed image in sRGB space.

To evaluate our approach, we require raw images of day and nighttime scenes taken with the same camera. Additionally, for quantitatively evaluating our method, and for training a baseline comparison model on purely real data, we require the nighttime images to have been captured in a paired manner with aligned short-exposure high-ISO noisy input images and long-exposure low-ISO noise-free ground truth images. To our knowledge, no such dataset exists in the literature. Note that we cannot use the SID~\cite{sid} dataset because it does not include day images. Therefore, we capture our own dataset to evaluate our proposed method. We use a Samsung S20 FE smartphone to capture data. Specifically, we capture images with the main 12 MP rear camera, which has a resolution of $4032 \times 3024$ pixels.

\begin{figure}[!b]
\setlength{\tabcolsep}{3pt}
\begin{tabular}{ccc}
\includegraphics[width=0.31\linewidth]{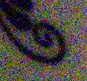}&
\includegraphics[width=0.31\linewidth]{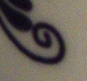}&
\includegraphics[width=0.31\linewidth]{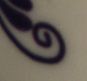}\\
Input ISO 3200 & ISO 50 & Average of \# 30 \\
single frame & single frame & ISO 50 frames
\end{tabular}
\caption{To generate the clean target, we average 30 long-exposure ISO 50 frames. This provides a better noise profile compared to using a single long-exposure ISO 50 frame. A high-ISO input frame is also shown for comparison.}
\label{fig:averaging}
\end{figure}

For nighttime scenes, we capture a sequence of images as follows. We first capture a burst of 30 frames at ISO 50, followed by two bursts of 10 frames at ISOs 1600, and 3200, respectively. We use the Android Camera2API to automate this capture sequence. For any given ISO, the exposure time is fixed for all frames. We select the exposure time corresponding to each ISO such that the image is correctly metered at 0 EV. The long-exposure ISO 50 images are used to generate the clean ground truth target, while the shorter-exposure higher-ISO frames are used as input. We image indoor and outdoor scenes of varying illuminance, such that the exposure time for the higher ISO input frames is a reasonably short value for hand-held imaging without camera shake.
However, note that we require aligned input and target frames for quantitative evaluation, and since the ISO 50 frames necessarily require a much longer exposure time, the camera and the scene are constrained to be static, similar to SID~\cite{sid}. To minimize camera motion during capture, the camera is mounted on a sturdy tripod, and the Camera 2 app is triggered remotely to avoid disturbing the camera while initiating the capture sequence.

\begin{figure*}[!t]
\centering
\includegraphics[width=1.0\linewidth]{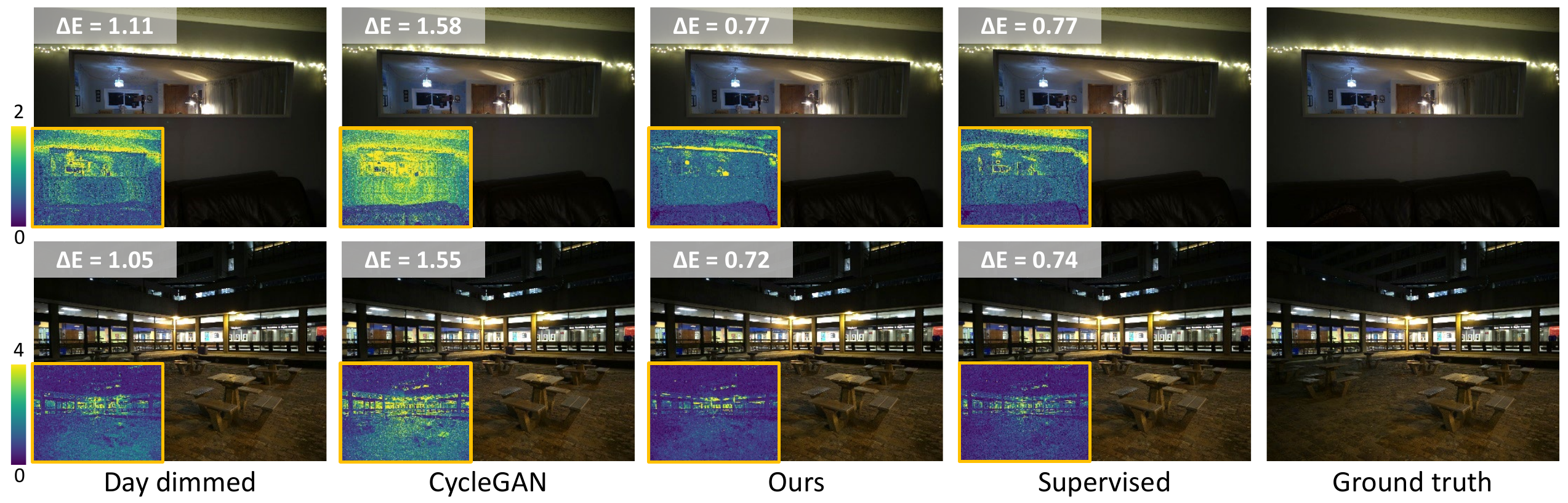}
\caption{Qualitative results for our neural ISP task assuming noise-free inputs. Inset shows $\Delta$E~\cite{deltaE} error map and average value.}
\label{fig:cr}
\end{figure*}

We image 105 nighttime scenes following the capture sequence described, for a total of 5250 nighttime raw images in our dataset. Our dataset contains a mix of indoor and outdoor scenes mimicking commonly encountered nighttime capture scenarios. In particular, the outdoor scenes are captured under street lighting, while the indoor scenes are captured under regular indoor illumination, such as incandescent lamps, florescent and LED lights. We also capture 70 day images at ISO 50. These images constitute the high-quality input data on which our day-to-night image synthesis algorithm is applied. Additionally, we capture 45 images of a gray card under different nighttime illuminations. These images are used to build the nighttime illumination dictionary that is used in the relighting step of our nighttime image synthesis algorithm. Representative examples from our dataset are shown in Fig. \ref{fig:dataset}.

\begin{figure*}[!t]
\centering
\includegraphics[width=1.0\linewidth]{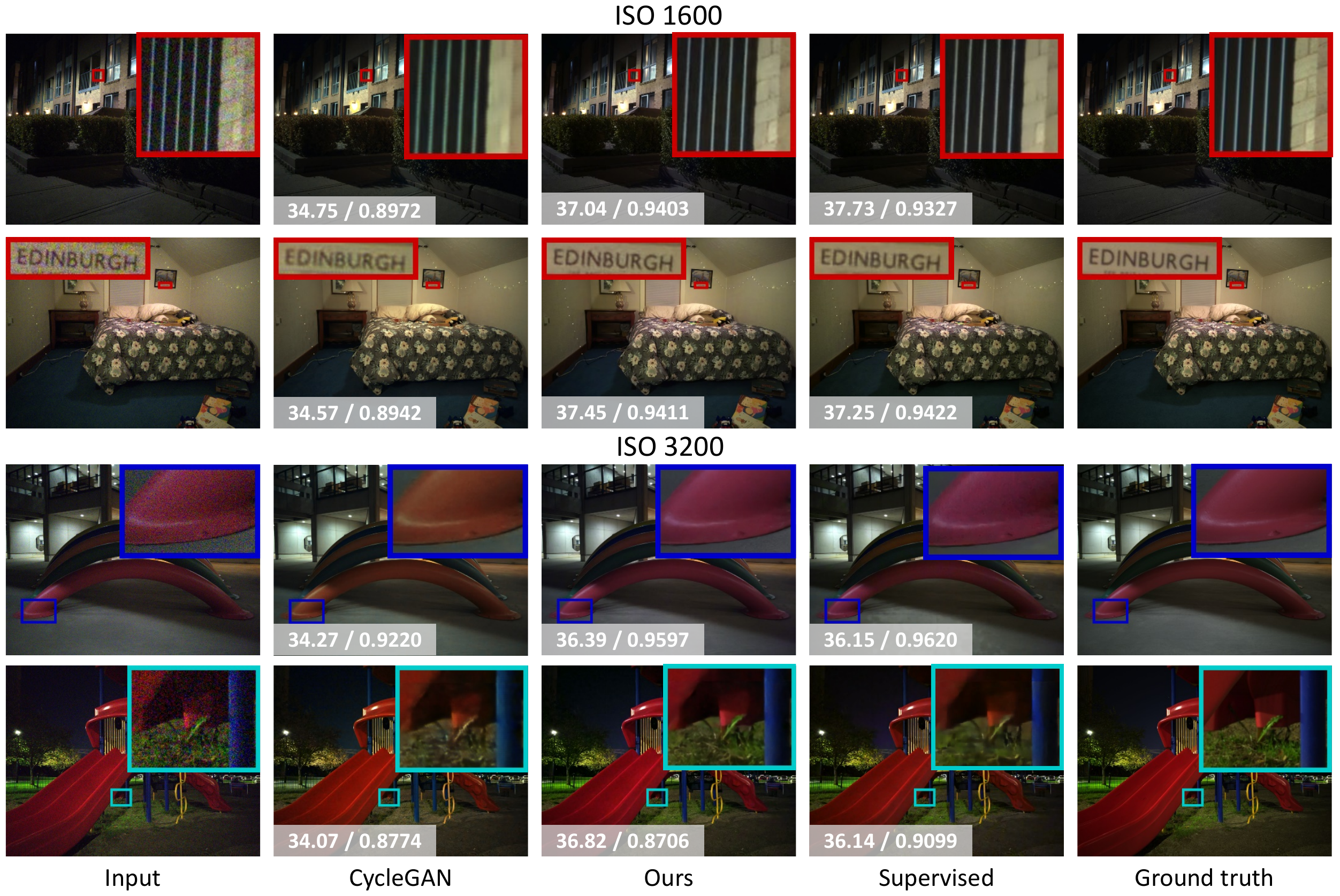}
\caption{Qualitative results for our neural ISP task with real noisy inputs. Inset shows zoomed-in region and PSNR (dB) / SSIM values.}
\label{fig:nr}
\end{figure*}

The SID dataset~\cite{sid} used a single long-exposure low-ISO image as the target. However, the SID dataset used DSLR cameras, while our data is collected using a smartphone camera that employs a smaller sensor with a poorer noise profile. Therefore, we process the nighttime bursts in our dataset as follows to generate the ground truth image. For each scene, we average the 30 long-exposure ISO 50 Bayer frames to produce a single merged Bayer frame. Since our images are carefully captured without any motion, we found that direct averaging without any further alignment gives good results. This averaged Bayer frame has a much more improved noise profile than any single frame from the burst. See Fig.~\ref{fig:averaging}. We then render this averaged Bayer frame through a software ISP~\cite{sidd} to produce the ground truth noise-free sRGB target image. Note that the averaged raw frame serves as the input for our initial noise-free neural ISP experiment. For the second neural ISP task, where we consider the effect of noise,  a single short-exposure high-ISO raw image is the input. While any of the 10 images in each of the ISO 1600 and ISO 3200 bursts could be used as input in this case, we use only the first frame for our experiments. However, note that the remaining frames are a valuable contribution as a dataset since each image is a distinct low-light image that can be used for training and testing. In Section~\ref{sec:isp} of supplementary, we also use these frames for a comparison against a burst denoising pipeline. In summary, we use 105 input/target pairs for training and evaluation for both our neural ISP experiments, without and with noise.

\subsection{Training}
We choose the standard UNet~\cite{unet} as our network architecture. UNet is well suited for image restoration problems, and in particular, its aptness as a nighttime neural ISP was demonstrated in SID~\cite{sid}. We train the network from scratch using the L1 loss and the Adam optimizer~\cite{Adam}. We crop non-overlapping patches of size $64 \times 64$ pixels. We train the network for 500 epochs with a batch size of 128 and an initial learning rate of $10^{-3}$, which is decayed to $10^{-4}$ after 400 epochs.

The target for the network is the processed sRGB image, while the input is a three-channel linear raw image that has been normalized, demosaiced, and white-balanced. The normalization step involves a simple black- and white-level adjustment and scaling. Demosaicing converts the single-channel raw Bayer frame to a three-channel RGB image. The white-balance operation performs a per-channel scaling of the RGB channels based on the estimated scene illumination. For our synthetic nighttime data, we set the white-balance vector to be $\frac{1}{N} \sum_{i=1}^{N} \mathbf{L}_{{\mbox{\scriptsize{night}}}_i}$ for both input and target during training. At test time, the white balance vector for the noisy input is extracted directly from the raw DNG file’s metadata. The ground truth at test time is rendered using the white balance vector corresponding to the first frame in the ISO 50 burst. In our experiments, we found that the camera ISP's AWB estimates are quite robust even in the presence of strong noise, such as at ISO 3200. To quantitatively verify this, we computed the angular error between the WB vector corresponding to the ISO 3200 input frame and the ISO 50 ground truth. We found the mean angular error over the 105 scenes in our dataset to be less than 1$^{\circ}$, which is considered visually imperceptible.
We adopt this strategy of applying the white balance before feeding the image to the network because we found it produces more accurate results than having the network learn to white-balance the image.


\section{Experiments}
\label{sec:expt}
We compare our day-to-night image synthesis with various baselines. Our preliminary baseline is to use day images directly for training. We also evaluate against a model that has only the dimming operation applied to the day images, without any relighting. A third baseline is to apply dimming followed by a \emph{global} relighting. In this case, a single illuminant is randomly sampled using our nighttime illuminant dictionary, and this illuminant is used to globally relight the image. Note that our proposed method applies \emph{local} relighting using a set of sampled illuminants. In addition to these three baselines that are derived from our proposed method, we compare against an unpaired image translation approach using CycleGAN~\cite{cyclegan}, and a supervised framework similar to SID~\cite{sid} using real paired data.

We first focus on the neural ISP task \emph{without} noise. To evaluate our proposed method, we partition the 70 daytime images into 60 images for training and 10 for validation. While the number of images is relatively small, these are full-resolution $4032 \times 3024$ images and provide sufficient data for patchwise training. We apply our image synthesis procedure of Section~\ref{sec:method} to generate our synthetic nighttime images, and we quantitatively evaluate our model trained on this data using the 105 real nighttime images. The peak-signal-to-noise (PSNR) and structural similarity index (SSIM)~\cite{SSIM} evaluated on the sRGB images are reported in Table~\ref{tab:color_rendering}. We also report the $\Delta$E value~\cite{deltaE}, which is widely used to measure changes in visual perception between two colors. A lower $\Delta$E is better. The three synthetic baselines described earlier are given the same training and validation split, and are trained using the same settings as our method.
Their results are also reported in the table, and it can be observed that our method outperforms these baselines.

\begin{table}[]
\centering
\caption{Quantitative results on our neural ISP task without noise. The models are partitioned based on whether the training data is synthetic only, a mix of synthetic and real, or purely real.}
\label{tab:color_rendering}
\resizebox{0.875\columnwidth}{!}{%
\begin{tabular}{llll}
\toprule
Model          & PSNR & SSIM  & $\Delta$E \\
\toprule
Day            & 41.16   & 0.9713 &  1.2369\\
Day dimmed     & 42.39   & 0.9832 & 1.0865\\
Global relight & 43.56   & 0.9860 & 0.9653\\
CycleGAN~\cite{cyclegan}       &          40.01 & 0.9675   & 1.6335    \\
Ours           & 45.28   & 0.9893 & 0.8759\\
\midrule
Ours 95\% + Real 5\%           &  45.99  & 0.9887 & 0.8682 \\
Ours 90\% + Real 10\%          &   46.01  & 0.9890  & 0.8559\\
\midrule

Supervised     & 46.19   & 0.9895 & 0.8605\\
\bottomrule
\end{tabular}%
}
\end{table}

We also perform a comparison where we replace our image synthesis algorithm with the well-established image-to-image translation method, CycleGAN~\cite{cyclegan}. In particular, we train CycleGAN on the unpaired day$\leftrightarrow$night task using our dataset of 70 daytime and 105 nighttime images. We use the noise-free average Bayer frame for the nighttime data. For both day and night data, we apply the white balance using the camera's AWB estimate.
We also demosaic the raw Bayer frame to a linear RGB image, to match the three-channel input used by CycleGAN. A random patch is cropped from every image at each training epoch. We use the official code and recommended hyper-parameters
to train the model. We then apply the trained model on the day images to translate them to night. We use this nighttime raw data, synthesized by CycleGAN, as input, and generate sRGB targets from them using the same software ISP pipeline~\cite{sidd} as applied for our method. We then train a UNet model for rendering with the same settings as used for training our model. The results on the 105 test images are reported in Table~\ref{tab:color_rendering}, and it can be observed that our method is more accurate even though CycleGAN's training had been exposed to the test data. Also note that a dataset of clean nighttime images is required for training CycleGAN, a limitation that our method explicitly addresses.

Our last comparison is against a supervised training setup using real paired data, similar to SID~\cite{sid}. They use a 4-channel stacked RGGB Bayer image at half the resolution as input to the UNet, and recover the full resolution at the output using a sub-pixel layer. Instead, we provide a demosaiced image as input to the UNet such that both input and output are 3-channel images at full resolution. We found this approach to be more accurate.
We use the 105 real nighttime pairs, and evaluate using three-fold cross validation. We use 60 images for training, and 10 for validation,  for a fair comparison with the synthetic data models. There are 35 testing images in each fold. Models are trained using the same UNet architecture and settings as our method, and the results are reported in the last row of Table~\ref{tab:color_rendering}. It can be observed that there is almost a 1 dB gap in performance between training on purely synthetic data using our algorithm and training exclusively on real images. To test our algorithm's utility as an augmentation strategy, we perform an experiment where we use a mix of synthetic and real data to train the model. Specifically, we use the same three-fold cross validation split, but the training data is comprised of $95\%$ synthetic data from our algorithm and $5\%$ real images. This allows us to narrow the 1 dB gap in performance to around 0.2 dB. We further tested a $90\%$ synthetic + $10\%$ real mixture that brings us even closer to the supervised model. Qualitative results of our method, along with comparisons, are shown in Fig.~\ref{fig:cr}.

\begin{table}[]
\centering
\caption{Quantitative results on our neural ISP task with real noisy inputs. The models are partitioned based on whether the training data is synthetic only, a mix of synthetic and real, or purely real.}
\label{tab:neural_isp}
\resizebox{\columnwidth}{!}{%
\begin{tabular}{lllll}
\toprule
\multirow{2}{*}{Model} & \multicolumn{2}{c}{ISO 1600} & \multicolumn{2}{c}{ISO 3200} \\
                       & PSNR       & SSIM        & PSNR       & SSIM        \\
\toprule
Day                    & 36.10        & 0.9215      & 33.55        & 0.8960      \\
Day dimmed             & 36.13        & 0.9254      & 33.47        & 0.8969      \\
Global relight         & 36.84        & 0.9353      & 35.63        & 0.9162      \\
CycleGAN               & 35.26               & 0.8968            &    33.58            &  0.8760           \\
Ours                   & 37.41        & 0.9368      & 35.70        & 0.9142      \\
\midrule
Ours 95\% + Real 5\%  & 38.32        & 0.9419      & 36.60        & 0.9206      \\

Ours 90\% + Real 10\%  & 39.04        & 0.9477      & 37.54        & 0.9352      \\
\midrule
Supervised             & 39.74        & 0.9541      & 38.16        & 0.9419 \\
\bottomrule
\end{tabular}%
}
\end{table}

We repeat these experiments for our second and more realistic task of a neural ISP that processes a \emph{noisy} raw input to sRGB. We used two target ISOs, 1600 and 3200 for testing. The results are reported in Table~\ref{tab:neural_isp}. For all the synthetic data, noise is synthetically added as described in Section~\ref{sec:method}. It can be observed that our proposed method performs better than CycleGAN and other baselines. Qualitative results are presented in Fig.~\ref{fig:nr}. We note that there is a gap in performance between our purely synthetic model  and the model in the last row trained exclusively on real data. While we selected the basic heteroscedastic noise model for simplicity, a complex noise simulator (e.g.,~\cite{wei2020physics,wei2021physics,abdelhamed2019noise}) that more realistically mimics the sensor noise could bridge this gap. However, we would like to highlight that with a mix of $10\%$ real data, the difference in performance is only a little over half a dB, as seen from Table~\ref{tab:neural_isp}. This experiment validates our method's effectiveness for data augmentation.


\section{Concluding remarks}

We have presented a procedure to convert daytime raw-RGB images to pairs of noisy/clean nighttime raw-sRGB images.  Our approach greatly reduces the time and effort required to prepare training data for neural ISPs targeting night mode imaging.  We note that a limitation of our work is that our selected software ISP~\cite{sidd} used to render target sRGB images does not include advanced photo-finishing routines such as local tone mapping.  The choice of the software ISP acts as an upper bound on the visual quality the trained neural ISP can produce. In addition, our selected high ISOs of 1600 and 3200 work well for typical nighttime scenes; however, we have not addressed extreme low light scenarios (e.g., moonlight) requiring much higher ISOs.

\newcommand{\hbAppendixPrefix}{S}
\renewcommand{\thefigure}{\hbAppendixPrefix\arabic{figure}}
\setcounter{figure}{0}
\renewcommand{\thetable}{\hbAppendixPrefix\arabic{table}}
\setcounter{table}{0}
\renewcommand{\theequation}{\hbAppendixPrefix\arabic{equation}}
\setcounter{equation}{0}
\renewcommand{\thesection}{\hbAppendixPrefix\arabic{section}}
\setcounter{section}{0}







\begin{figure}[!b]
\setlength{\tabcolsep}{3pt}
\begin{tabular}{c}
\includegraphics[width=\columnwidth]{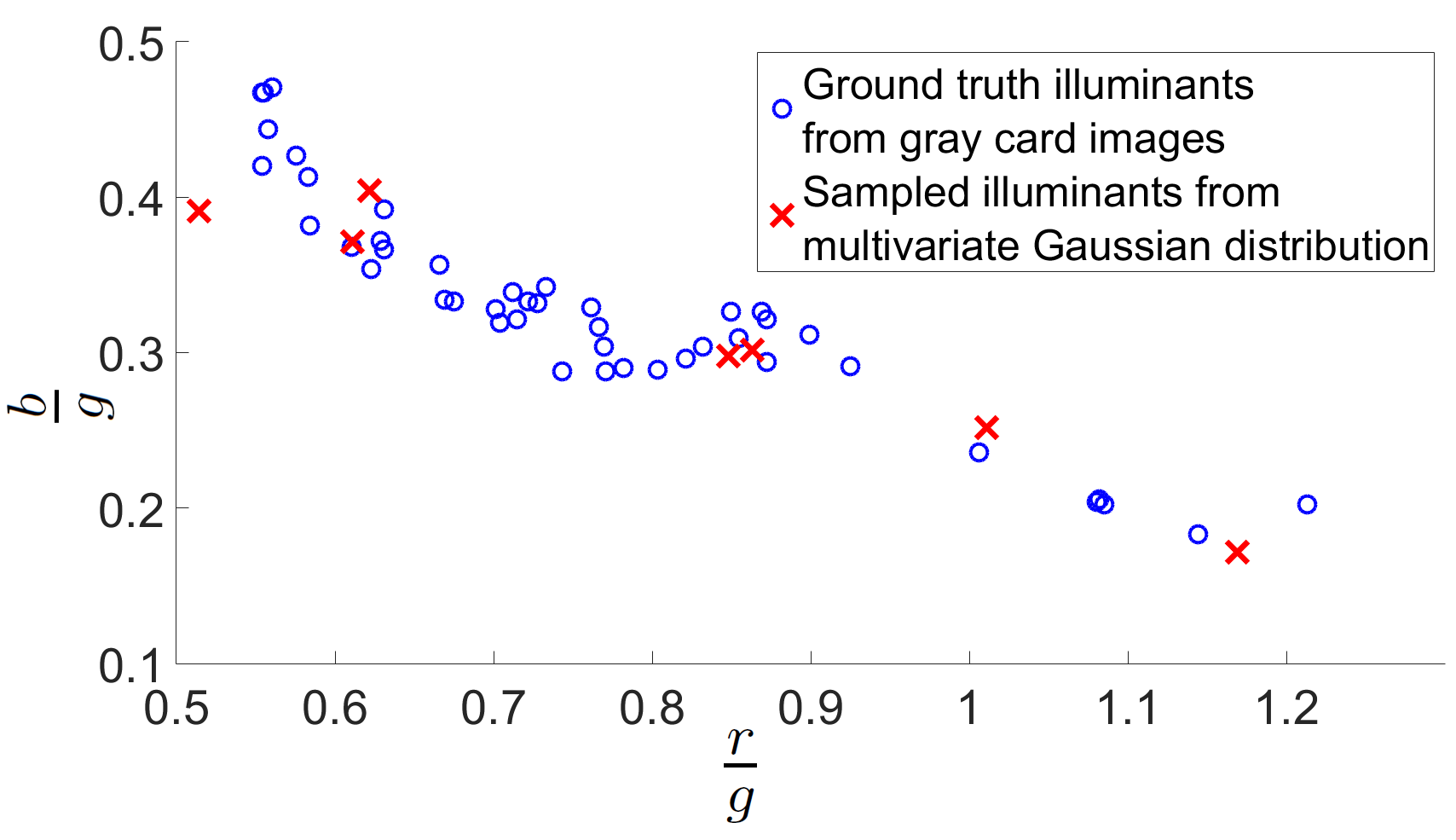}
\end{tabular}
\caption{To relight the scene using our proposed day-to-night image synthesis framework, we draw random samples (red markers) from a 2D multivariate Gaussian distribution of joint $[\frac{r}{g},\frac{b}{g}]$ chromaticity values fit on a database of night illuminations (blue markers) measured using a gray card.}
\label{fig:illum}
\end{figure}

\begin{figure*}[!t]
\centering
\includegraphics[width=1.0\linewidth]{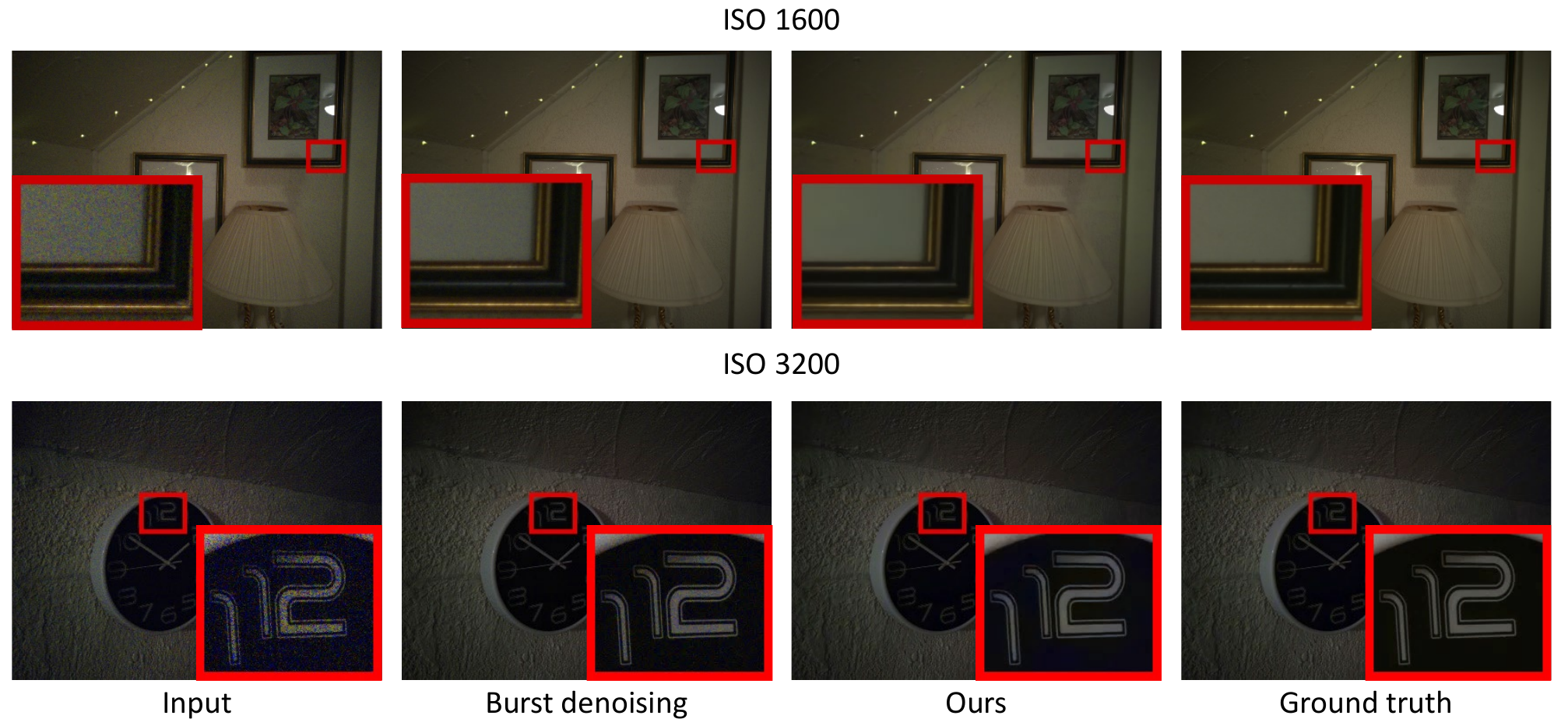}
\caption{Comparison to burst denoising.}
\label{fig:denoising}
\end{figure*}

\section{Supplementary material}
\subsection{Illumination sampling}
\label{sec:illum}

One of the main steps of our day-to-night image synthesis framework is to relight the image with night illuminants as explained in Section~\ref{sec:method}. To this end, we first construct a night time illuminant dictionary $\mathcal{L}$ by imaging gray cards under different nighttime illuminations. The blue markers in the plot of Fig.~\ref{fig:illum} show the $[\frac{r}{g},\frac{b}{g}]$ chromaticity values of these ground truth illuminations recorded by the gray card. To apply the relighting step of our pipeline, we first fit a 2D multivariate Gaussian distribution of joint chromaticity values around our database of night illuminations $\mathcal{L}$, and then sample from this distribution, as described in Equations~\ref{eqn1} and~\ref{eqn2}. The red markers in the plot show a few such randomly drawn samples. We use these samples to locally relight the scene and generate a synthetic nighttime image.

\subsection{Comparison to burst denoising}
\label{sec:isp}
For each scene in our nighttime dataset, we have a burst of 10 frames each at ISOs 1600 and 3200. Our experiments in Section~\ref{sec:expt} were performed using only the first image in the high ISO bursts. Here, we perform a comparison to a burst denoising pipeline using all 10 frames. Since our images are already aligned, we directly average the Bayer frames. Then, we render the averaged Bayer image through the software ISP in~\cite{sidd}. Note that this is an idealized burst denoising pipeline that benefits from perfect alignment, which is not available in practice. Quantitative results are presented in Table~\ref{tab:burst_denoising}. For ease of comparison, the results of our method are reproduced from Table~\ref{tab:neural_isp}. It can be observed that we outperform the burst denoising pipeline by a sound margin. Qualitative comparisons are provided in Fig.~\ref{fig:denoising}. 

\begin{table}[!h]
\centering
\caption{Comparison to burst denoising.}
\label{tab:burst_denoising}
\resizebox{0.9\columnwidth}{!}{%
\begin{tabular}{lllll}
\toprule
\multirow{2}{*}{Method} & \multicolumn{2}{c}{ISO 1600} & \multicolumn{2}{c}{ISO 3200} \\
                       & PSNR       & SSIM        & PSNR       & SSIM        \\
\toprule
Burst denoising                    & 34.91        & 0.8021      & 31.39        & 0.6526      \\
Ours                   & 37.41        & 0.9368      & 35.70        & 0.9142      \\
\bottomrule
\end{tabular}%
}
\end{table}

\subsection{Additional qualitative results}
\label{sec:qualitative}

\begin{figure*}[!t]
\centering
\includegraphics[width=1.0\linewidth]{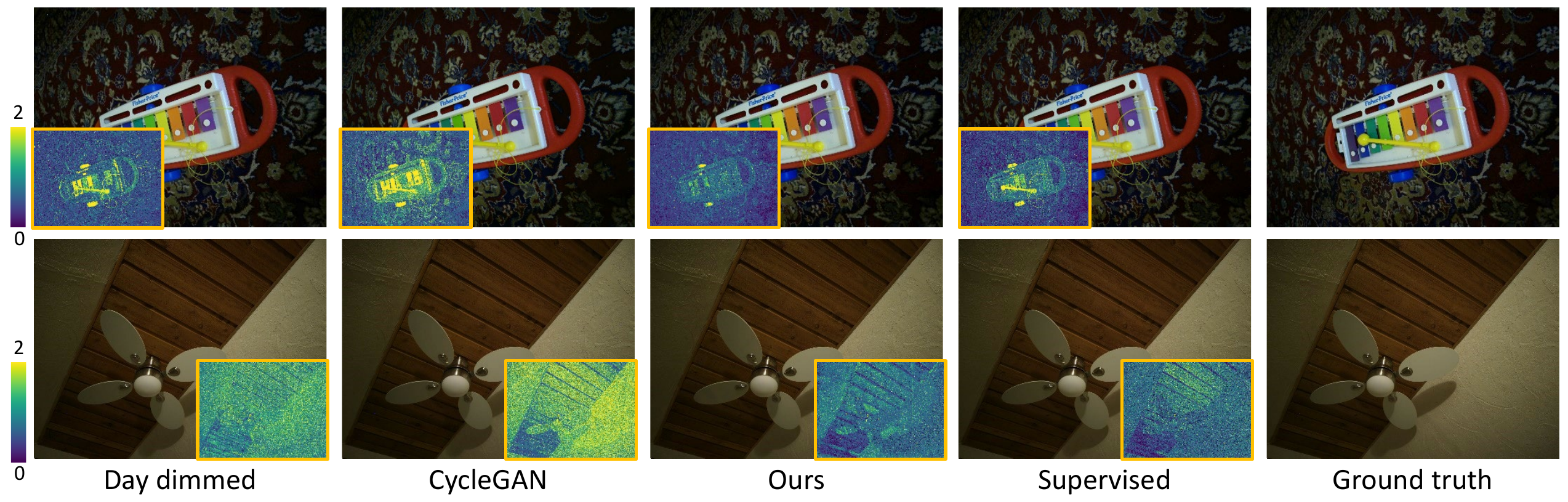}
\caption{Additional qualitative results for our neural ISP task assuming noise-free inputs. Inset shows $\Delta$E~\cite{deltaE} error map.}
\label{fig:cr1}
\end{figure*}

\begin{figure*}[!t]
\centering
\includegraphics[width=1.0\linewidth]{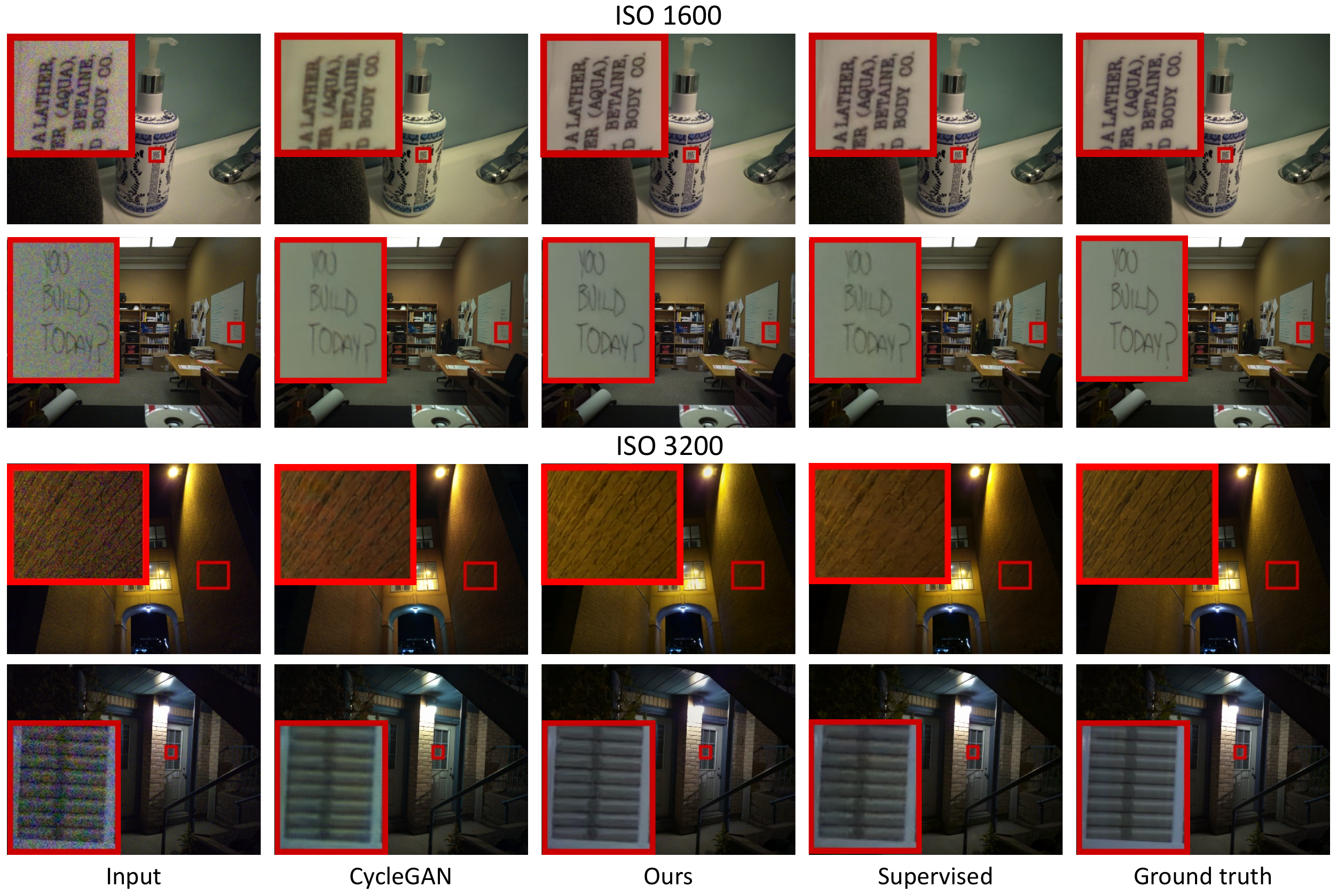}
\caption{Additional qualitative results for our neural ISP task with real noisy inputs. Inset shows zoomed-in regions.}
\label{fig:nr1}
\end{figure*}

We provide additional qualitative results of our neural ISP tasks without and with noise in Fig.~\ref{fig:cr1} and Fig.~\ref{fig:nr1}. These figures extend Fig.~\ref{fig:cr} and Fig.~\ref{fig:nr}, respectively.

\subsection{Ablation on the loss function}
A comparison between L2 and L1 loss (proposed) is presented in Table~\ref{tab:loss} for our day-to-night model. PSNR (dB) and SSIM values are reported. L1 loss results have been reproduced from Table~\ref{tab:color_rendering} (No noise) and Table~\ref{tab:neural_isp} (ISO 1600/3200). As seen from the results, the proposed L1 loss is generally more accurate than L2.
\begin{table}[!h]
\centering
\caption{An ablation on the loss function.}
\label{tab:loss}
\resizebox{0.875\columnwidth}{!}{%
\begin{tabular}{|c|c|c|c|}
\hline
Loss & No noise & ISO 1600 & ISO 3200 \\
 \hline
L2  & 45.15 / 0.9887    &  37.10 / 0.9348    & 35.41 / \textbf{0.9157} \\
\hline
L1  & \textbf{45.28} / \textbf{0.9893}    &  \textbf{37.41} / \textbf{0.9368}    & \textbf{35.70} / 0.9142 \\
\hline
\end{tabular}
}
\end{table}

\subsection{Comparison with EnlightenGAN}
In Section~\ref{sec:expt}, we had compared our method to CycleGAN~\cite{cyclegan}. In Table~\ref{tab:enlighten}, we also report the results of EnlightenGAN~\cite{jiang2021enlightengan}, another popular image-to-image translation technique. We used the same training setup as used for CycleGAN. PSNR (dB) and SSIM values are reported. Our results have been reproduced from Table~\ref{tab:color_rendering} (No noise) and Table~\ref{tab:neural_isp} (ISO 1600/3200). It can be observed that we outperform EnlightenGAN by a sound margin. 
\begin{table}[!h]
\caption{Comparison with EnlightenGAN~\cite{jiang2021enlightengan}.}
\label{tab:enlighten}
\centering
\resizebox{\columnwidth}{!}{%
\begin{tabular}{|c|c|c|c|}
\hline
Method & No noise & ISO 1600 & ISO 3200 \\
 \hline
EnlightenGAN  &  38.95 / 0.9652   &  35.24 / 0.9203    & 32.63 / 0.8879 \\
\hline
Ours  & \textbf{45.28 / 0.9893}    &  \textbf{37.41 / 0.9368}    & \textbf{35.70 / 0.9142} \\
\hline
\end{tabular}
}
\end{table}


{\small
\bibliographystyle{ieee_fullname}
\bibliography{egbib}
}

\end{document}